\documentclass[pdflatex]{sn-jnl}
\usepackage{natbib}
\usepackage{graphicx}%
\usepackage{multirow}%
\usepackage{amsmath,amssymb,amsfonts}%
\usepackage{amsthm}%
\usepackage{mathrsfs}%
\usepackage[title]{appendix}%
\usepackage{xcolor}%
\usepackage{textcomp}%
\usepackage{manyfoot}%
\usepackage{booktabs}%
\usepackage{algorithm}%
\usepackage{algorithmicx}%
\usepackage{algpseudocode}%
\usepackage{listings}%
\usepackage{subfigure}
\usepackage{wrapfig}

\newcommand{\name}{{\sc Apart }}
\newcommand{\namea}{{\sc Apart}}

\begin{document}

\title[Adaptive Adapter Routing for Long-Tailed Class-Incremental Learning]{Adaptive Adapter Routing for Long-Tailed Class-Incremental Learning}

\author[1,2]{\fnm{Zhi-Hong} \sur{Qi}}\email{qizh@lamda.nju.edu.cn}

\author*[1,2]{\fnm{Da-Wei} \sur{Zhou}}\email{zhoudw@lamda.nju.edu.cn}

\author[1]{\fnm{Yiran} \sur{Yao\footnotemark[3] \ }}\email{yrharryyao@gmail.com}

\author[1,2]{\fnm{Han-Jia} \sur{Ye}}\email{yehj@lamda.nju.edu.cn}

\author[1,2]{\fnm{De-Chuan} \sur{Zhan}}\email{zhandc@nju.edu.cn}

\affil[1]{\orgdiv{School of Artificial Intelligence}, \orgname{Nanjing University}, \orgaddress{\street{163 Xianlin Avenue}, \city{Nanjing}, \postcode{210023}, \state{Jiangsu}, \country{China}}}

\affil[2]{\orgdiv{National Key Laboratory for Novel Software Technology}, \orgname{Nanjing University}, \orgaddress{\street{163 Xianlin Avenue}, \city{Nanjing}, \postcode{210023}, \state{Jiangsu}, \country{China}}}

\renewcommand{\thefootnote}{\fnsymbol{footnote}}
\footnotetext[1]{Correspondence to: Da-Wei Zhou (zhoudw@lamda.nju.edu.cn)}
\footnotetext[3]{Work done during an internship at School of Artificial Intelligence, Nanjing University.}
\renewcommand{\thefootnote}{\arabic{footnote}}

\abstract{In our ever-evolving world, new data exhibits a long-tailed distribution, such as emerging images in varying amounts. This necessitates continuous model learning imbalanced data without forgetting, addressing the challenge of long-tailed class-incremental learning (LTCIL). Existing methods often rely on retraining linear classifiers with former data, which is impractical in real-world settings. In this paper, we harness the potent representation capabilities of pre-trained models and introduce AdaPtive Adapter RouTing (\namea) as an exemplar-free solution for LTCIL. To counteract forgetting, we train inserted adapters with frozen pre-trained weights for deeper adaptation and maintain a pool of adapters for selection during sequential model updates. Additionally, we present an auxiliary adapter pool designed for effective generalization, especially on minority classes. Adaptive instance routing across these pools captures crucial correlations, facilitating a comprehensive representation of all classes. Consequently, \name tackles the imbalance problem as well as catastrophic forgetting in a unified framework. Extensive benchmark experiments validate the effectiveness of \namea. Code is available at: \href{https://github.com/vita-qzh/APART}{https://github.com/vita-qzh/APART}.}


\maketitle


\section{Introduction}
Traditional machine learning algorithms typically assume a closed-world scenario, where data originates from a static, balanced distribution~\cite{ye2024contextualizing}. However, in reality, data often exhibits a long-tailed streaming pattern, as pictures emerge all the time but are different in category and number. This necessitates incremental learning from long-tailed data, referred to as Long-Tailed Class-Incremental Learning (LTCIL)~\cite{liu2022long}.
In LTCIL, a significant challenge is catastrophic forgetting~\cite{french1999catastrophic}, where the model tends to lose knowledge of former data during the learning process. Additionally, the inherent data imbalance causes the model to under-represent minority classes, leading to bias towards majority classes~\cite{zhang2023deep}. 
These interrelated challenges pose a significant problem in the machine learning community. Consequently, several algorithms have been developed to address them. For instance, LWS~\cite{liu2022long} rebalances the classifier by sampling a balanced dataset from both new and reserved former data. Similarly, GVAlign~\cite{Kalla_2024_WACV} enhances representation robustness and aligns the classifier by replaying generated pseudo-augmented samples. These approaches yield improved performance but rely on the storage of \emph{exemplars} from former classes.

Retaining exemplars is critical to preventing forgetting in LTCIL, but this approach often fails in real-world applications due to storage limitations~\cite{krempl2014open} and privacy concerns~\cite{chamikara2018efficient}. Recent advancements in Pre-Trained Models (PTMs)~\cite{han2021pre}, however, demonstrate their effectiveness without relying on exemplars, thanks to their robust representations. PTMs are increasingly favored not only in class-incremental learning~\cite{l2p} but also in long-tailed learning~\cite{pel}, challenging the traditional `training-from-scratch' paradigm. Their ability to provide strong foundational knowledge, enhanced by extensive pre-training datasets, ensures impressive adaptability to downstream tasks. This adaptability proves particularly beneficial in handling data scarcity in minority classes and maintaining performance on older tasks. As such, leveraging PTMs has become a prominent strategy in these fields to achieve superior performance. In this paper, we explore the integration of PTMs into LTCIL, aiming to overcome its challenges in an `{\em exemplar-free}' manner.

LTCIL presents two critical challenges, {\em i.e.,} catastrophic forgetting and data imbalance. Catastrophic forgetting occurs when new information supersedes old knowledge, leading to the overwriting of existing features and overall performance decay. Data imbalance, on the other hand, skews the learning process towards the majority class, neglecting the minority class. As a result, the boundary between the majority and minority classes is biased, making the model more likely to classify samples into the majority classes. In the learning process, these two issues are closely coupled, amplifying the difficulty of LTCIL.

An ideal model capable of continuously learning from a long-tailed data stream should be both \emph{provident and comprehensive}. Provident means the model, especially when based upon PTMs, can fully exploit its potential to learn new classes. Consequently, the model updating mechanism should be carefully designed to resist catastrophic forgetting in the learning process. Meanwhile, comprehensive implies that the model employs a specific learning strategy to capture minority class more, offering a holistic and distinct representation for all. 
By harmonizing these provident and comprehensive aspects, a model can adeptly navigate the challenges of LTCIL, paving the way for enhanced performance.

In this paper, we propose AdaPtive Adapter RouTing (\namea) to address the above challenges in LTCIL. To make the model provident, we freeze majority of the parameters of PTMs and employ trainable adapters at each layer. Furthermore, we extend one group of adapters to a pool containing multiple groups. Every time new data comes, we retrieve the most relevant group of adapters and update it. Additionally, we introduce an auxiliary pool specifically focused on learning from minority classes. During inference, we 
dynamically combine these two pools to get a comprehensive representation. Rather than using a fixed threshold to filter training data for the auxiliary pool, our method adaptively learns instance routing, encoding task information in a data-driven manner. This reflects the auxiliary pool's relevance to minority classes, enabling a holistic overview of all classes. The unified framework above is trained in an end-to-end fashion, enabling automatic routing learning in a data-driven way. We extensively validate the effectiveness of \name through numerous experiments on several benchmark datasets.

The main contributions of \name can be summarized as follows:
\begin{itemize}
\item Layer-wise adapters for deeper adaptation of pre-trained models are selected among multiple alternatives, reducing forgetting when transferring to downstream tasks.
\item An auxiliary pool is specially designed for minority classes to compensate for the lack of data. The imbalance in data when training the auxiliary pool greatly decreases, resulting in a comprehensive representation of minority classes.
\item Adaptive routing is learned to capture correlations between data and the auxiliary pool automatically in a data-driven manner, reducing dependency on the manual threshold when defining minority classes.
\end{itemize}
This paper is organized as follows. Section \ref{related} reviews the main related work. Section \ref{formulation} formulates the investigated issue and introduces the baseline method. Section \ref{method} describes \name and details each of its elements. Section \ref{exp} presents the empirical evaluations and further analysis. After that, we conclude the paper in Section \ref{conclusion}.

\section{Related Work}\label{related}

\noindent\textbf{Long-Tailed Learning:} aims to learn from highly imbalanced data~\cite{zhang2023deep}, where a small number of classes (majority classes/head classes) have a large amount of data, while rest classes (minority classes/tail classes) have limited data. 
Current algorithms can be roughly divided into three groups. The first group considers re-sampling the dataset to form a balanced training set~\cite{smote} or re-weighting the loss terms to favor tail classes~\cite{banlancedloss,focal}. The second group considers transferring knowledge from head classes to tail classes~\cite{wang2017learning} and self-training~\cite{rosenberg2005semi,wei2021robust} to enhance the recognition ability of tail classes. The third group designs techniques to improve representation or classifier modules via representation learning~\cite{huang2016learning}, decoupled training~\cite{kang2020decoupling}, and ensemble learning~\cite{zhou2020bbn}. 

\noindent\textbf{Class-Incremental Learning (CIL):} aims to sequentially learn new tasks without forgetting former knowledge~\cite{zhou2024class,wang2023few}. To alleviate the dilemma, a large number of work is proposed, mainly falling into three categories. The first group transfers the knowledge of old models to the new one by knowledge distillation~\cite{hinton2015distilling} when updating~\cite{li2017learning,douillard2020podnet}. The second group is based on the reserved exemplars of former tasks and replays them to maintain old knowledge~\cite{zhengmulti,rebuffi2017icarl,hou2019lucir,castro2018eeil}. The third group expands the network~\cite{yan2021der,wang2022foster,zhou2022memo} to meet the demand for model capacity arising from the increasing data. As pre-trained models gain popularity, more methods based on PTMs emerge~\cite{zhou2024class,l2p,seale2022coda,zhou2024continual,wang2022dualprompt,zhou2024expandable}. These methods mainly design lightweight modules to adapt the PTM in a parameter-efficient manner.

\noindent\textbf {Long-Tailed Class-Incremental Learning (LTCIL):} is recently proposed to learn from long-tailed streaming data. LWS~\cite{liu2022long} samples a balanced dataset from new data and reserved former data to re-weight the classifier for better performance. Furthermore, GVAlign~\cite{Kalla_2024_WACV} enhances the robustness of representations and aligns the classifier by replaying generated pseudo-augmented samples. These methods both follow a two-stage strategy, rectifying the outputs of the model using a balanced dataset in the second stage. In contrast, we aim for a better incremental performance without accessing former data.

\section{Preliminaries}\label{formulation}
In this section, we first describe the setting of LTCIL and then introduce the baseline method and its limitations.

\subsection{Long-Tailed Class-Incremental Learning}
\label{ltcil}
In class-incremental learning, a model learns from sequential tasks. When task $t$ arrives, the training dataset for the model is denoted as $D_t = \{(\mathbf{x}_i, y_i)\}_{i=1}^{n_t}$, where $\mathbf{x}_i \in \mathbb{R}^{H \times W \times C}$ is the $i$-th instance, $y_i \in \mathbf{Y}_t$ is the corresponding label in current label space and $n_t$ is the total size of task $t$. There are no overlaps in labels, \emph{i.e.}, $\mathbf{Y}_t \bigcap \mathbf{Y}_{t'} = \emptyset$, when $t \neq {t'}$. Different from the uniform distribution where the frequency of each class is a constant in the conventional CIL, data in LTCIL follows a long-tailed distribution.
The steeper the distribution is, the more challenging the problem is for the model to fit the tail classes without forgetting.

In LTCIL, we denote the classification model as $f:\mathbb{R}^{H \times W \times C} \rightarrow \mathbb{R}^{|\mathcal{Y}_t|}$, where $\mathcal{Y}_t = \bigcup_{k=1}^t \mathbf{Y}_k$ is the set of all seen classes at task $t$. It can be decoupled as a feature extractor $\phi:\mathbb{R}^{H \times W \times C} \rightarrow \mathbb{R}^d$ ($d$ is the dimension of the feature) and a classifier $g:\mathbb{R}^d \rightarrow \mathbb{R}^{|\mathcal{Y}_t |}$, \emph{i.e.}, $f(\mathbf{x}) = g(\phi(\mathbf{x}))$. The classifier can be decomposed as a set of classifiers for each task, \emph{i.e.}, $g=[g_1, \dots, g_t]$, where $g_k:\mathbb{R}^d \rightarrow \mathbb{R}^{|\mathbf{Y}_k|}$. When facing a new task, the classifier $g$ needs to be updated and extended. 
We follow~\cite{zhu2021self,wang2022dualprompt} to implement all methods \emph{without exemplars}, \emph{i.e.}, when training on task $t$, the model only has access to $D_t$. As LTCIL provides no task id at inference, the model must distinguish between old and new classes and have a good performance on all seen tasks. In other words, the performance on the whole balanced testing dataset $\bigcup_{k=1}^t D_k^{test}$ is taken into consideration, \emph{i.e.}, $\sum_{(\mathbf{x}_i, y_i) \in \bigcup_{k=1}^t D_k^{test}} \ell(f(\mathbf{x}_i), y_i)$, where $\ell(\cdot,\cdot)$ measures the discrepancy between the input pairs.

\subsection{Pre-Trained Models for CIL}

Currently, pre-trained models gain popularity in the CIL field. The most representative PTM for visual recognition tasks is Vision Transformer (ViT)~\cite{dosovitskiy2020image} pre-trained on the large-scale dataset (\emph{e.g.,} ImageNet~\cite{russakovsky2015imagenet}) as the backbone $\phi$ to extract features of the instances. ViT consists of an embedding layer and several transformer blocks. An image $\mathbf{x}$ is firstly divided into a sequence of patches and then passes the embedding layer to get its embedding $\mathbf{E}=[\mathbf{e}_0, \mathbf{e}_1, \dots, \mathbf{e}_{N_p}] \in \mathbb{R}^{N_p \times d}$ ($N_p$ is the number of patches). Then a learnable \textbf{[CLS]} token $\mathbf{c} \in \mathbb{R}^{d}$ is added to its embedding to get the final input of transformer blocks $\mathbf{x}_0 = [\mathbf{c}, \mathbf{E}]$. The model gives the ultimate prediction based on $\phi(\mathbf{x})$, the embedded feature of \textbf{[CLS]} token.

\begin{figure*}[t]
	\vspace{-2mm}
	\begin{center}
		{\includegraphics[width=\columnwidth ]{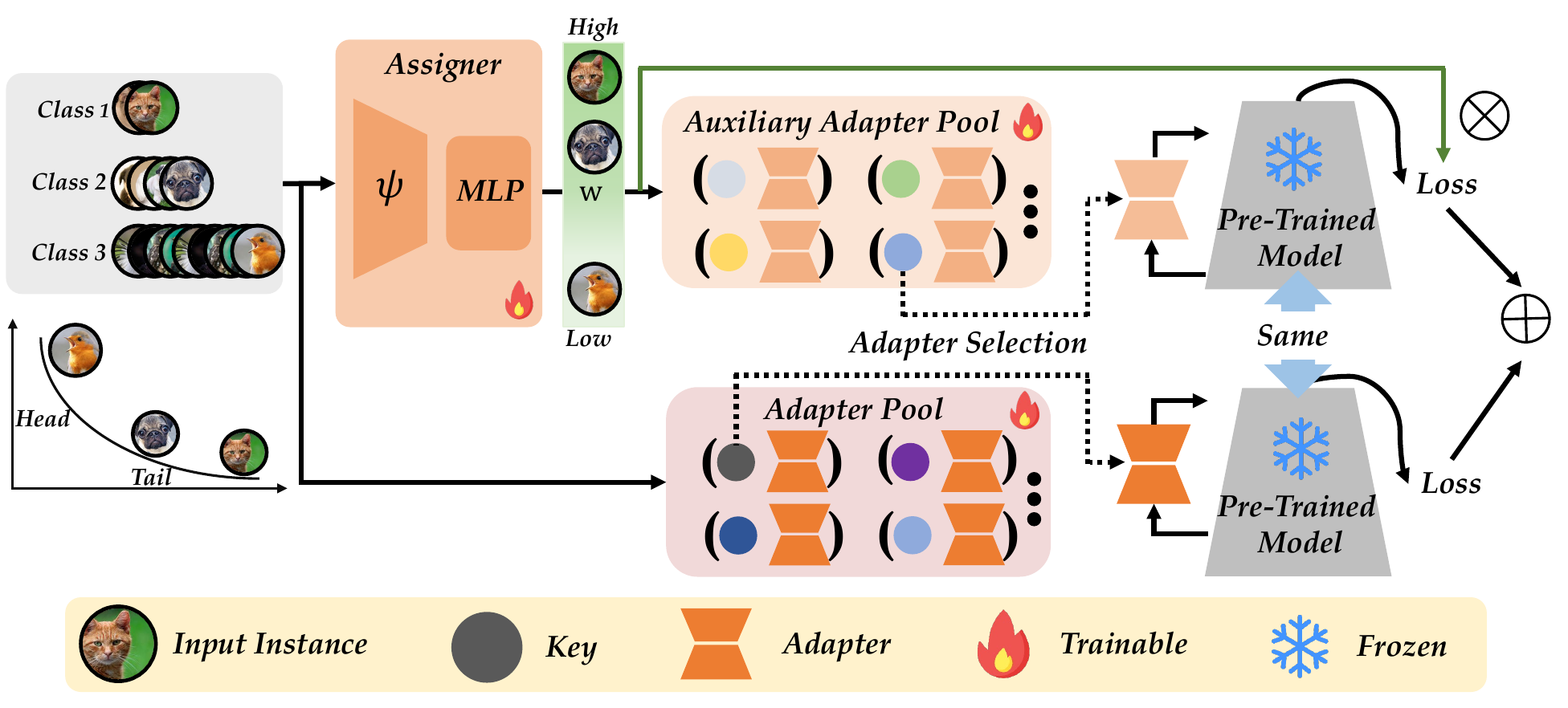}
		}
	\end{center}
 \vspace{-2mm}
	\caption{ Demonstration of \namea. To make the model comprehensive, an auxiliary adapter pool is learned for minority classes and instance-wise routing is learned adaptively. To make the model provident, multiple adapters form a pool to enlarge the capacity of fine-tuned model. The objective is to learn an automatic routing to learn effectively from minority classes without forgetting.  
	\vspace{-2mm}
	} \label{figure:method}
\end{figure*}

L2P~\cite{l2p} is the first method to utilize pre-trained ViT in CIL. To adapt to the downstream tasks efficiently, it employs visual prompt tuning~\cite{jia2022visual}, a parameter-efficient fine-tuning technique. Prompts can be seen as a horizontal expansion of the input. During training, it freezes the whole backbone and prepends prompt $\mathbf{P} \in \mathbb{R}^{L \times d}$ to the embedding of the instance, where $L$ is the length of the prompt. Then, the concatenated embedding $[\mathbf{c}, \mathbf{P}, \mathbf{E}]$ is passed to the frozen backbone to get the features for classification. In this way, knowledge about the task is encoded in the prompts. To mitigate forgetting, multiple prompts are provided for the model to enlarge the representation space, denoted as a pool $\mathcal{P} = [\mathbf{P}_1, \mathbf{P}_2, \dots, \mathbf{P}_M]$. For each prompt $\mathbf{P}_i$, a key $\mathbf{k}_i$ is associated in the key-value format for query ({\em i.e.,} $(\mathbf{k}_i, \mathbf{P}_i)$). At training, instance-wise prompt is chosen from $\mathcal{P}$ and then updated. The choice of prompt is based on the distance between the instance and the learnable keys: 
\begin{align}
\label{eq:choose}
\min_s \gamma(\phi(\mathbf{x}), \mathbf{k}_s) \,,
\end{align}
where $\gamma(\cdot, \cdot)$ denotes cosine distance. Here, we use the classification feature $\phi(\mathbf{x})$ to represent the instance, and calculate the distance with $\mathbf{k}_s$, the key for $\mathbf{P}_s$. Other prompt-based algorithms~\cite{wang2022dualprompt,seale2022coda} also explore more choices and fusion mechanisms.

\noindent{\textbf{Discussions:}} While Eq.~\ref{eq:choose} provides an efficient way to encode task-specific information in the prompts, L2P does not obtain a promising performance in LTCIL. There are two reasons accounting for the degradation. Firstly, learnable prompts are only prepended to the input level and thus have limited influence when the whole backbone is frozen. It restricts the model's representation ability when facing diverse downstream tasks. Secondly, when the data distribution becomes long-tailed, it requires more specific measures to compensate for the minority classes for a holistic feature. During training, prompts are unavoidably biased towards the majority classes and can not store as precise knowledge as expected. Thus, the ability to resist forgetting of Eq.~\ref{eq:choose} is weakened.


\section{Adaptive Adapter Routing for LTCIL}\label{method}

Motivated by the potential of PTMs, we try to incorporate PTMs into LTCIL in an exemplar-free manner. To make the model provident, we train adapters inserted at each layer instead of prepending prompts at the first layer, vertically expanding the frozen PTM and making a deeper adaptation. On the other hand, to make the model comprehensive, an additional mechanism unique for minority classes is proposed. To strengthen the correlation of the addition and minority classes, instance routing is adaptively learned in a data-driven manner. In this section, we first introduce the techniques to facilitate long-tailed learning, and then discuss the routing strategy. We summarize the training pipeline in the last part.

\subsection{Auxiliary Adapter Pool}

Adapter is a vertical expansion of the PTMs, increasing the transferability to downstream tasks while consuming limited parameters. Compared to prompts, the adaptation occurs in the structure instead of the input, resulting in a better performance on visual recognition tasks. In this paper, we follow~\cite{chenadaptformer} to insert the adapter to each transformer block in ViT. Adapter is a bottleneck structure consisting of a down-projection layer $W_{down} \in  \mathbb{R} ^{d \times r}$, a non-linear activation $\textbf{ReLU}$ and an up-projection layer $W_{up} \in \mathbb{R} ^{r \times d}$, where $r \ll d$ is the bottleneck middle dimension. It mainly changes the residual connection in the transformer blocks by adding non-linear transformations to the identical input. 
For the $i$-th transformer block, we denote its output after the multi-head self-attention as $\hat{\mathbf{x}}_i$. We insert adapters to the MLP structure and get the output as:
\begin{equation}
\begin{aligned}
\mathbf{x}_{i+1} = \textbf{MLP}(\textbf{LN}(\hat{\mathbf{x}}_i)) +  \textbf{ReLU}(\hat{\mathbf{x}}_iW_{down})W_{up} \,,
\end{aligned}
\end{equation}
where $\textbf{MLP}(\textbf{LN}(\hat{\mathbf{x}}_i))$ is the original output of the transformer block, and $\textbf{LN}$ denotes layer norm.
We denote the group of adapters inserted at each block as $\mathbf{A}$. 
During training, we freeze the whole backbone and only optimize the lightweight modules in $\mathbf{A}$, enabling the adaptation to downstream tasks while preserving PTM's representations.

To enlarge the capacity of the fine-tuned model and mitigate forgetting, we  follow~\cite{l2p} to expand adapters $\mathbf{A}$ to a pool $\mathcal{A}$, which contains $M$ groups of adapters $\mathcal{A} = [\mathbf{A}_1, \mathbf{A}_2, \dots, \mathbf{A}_M]$. 
We also adopt the key-query matching strategy in Eq.~\ref{eq:choose}. Different from L2P, which selects a set of prompts, \name chooses only one group of adapters, \emph{i.e.}, 12 adapters inserted into each layer, which is enough to store exacted knowledge. Then, the adapted model gives the prediction
\begin{equation*}
    f(\mathbf{x}; \mathcal{A}) = g(\phi(\mathbf{x};\mathcal{A}))
\end{equation*}
where $\phi(\cdot;\mathcal{A})$ is the adapted feature extractor based on the pool $\mathcal{A}$ and $g(\cdot)$ is the corresponding tuned classifier which is completed with a simple linear layer.

During training, we seek to find the most suitable adapters within $\mathcal{A}$ to adapt to the current task by minimizing:
\begin{equation}
\label{pool}
\begin{aligned}
 \mathcal{L}(\mathbf{x}, y; \mathcal{A}) & = \ell(f(\mathbf{x};\mathcal{A}), y) + \gamma(\phi(\mathbf{x}), \mathbf{k}_s) \,,
\end{aligned}
\end{equation}
where $\mathbf{k}_s$ is the key of the most suitable group chosen from $\mathcal{A}$ according to Eq. \ref{eq:choose}. Solving Eq.~\ref{pool} enables encoding the task-specific information into the group of adapters, and the adapter selecting strategy enables a holistic estimation of the query instance to the most suitable adapter.

\noindent\textbf{Auxilliary Adapter Pool}: Since imbalanced distribution makes a biased model and insufficient learning from minority classes, to compensate for the shortfall, an intuitive approach is to specifically train minority classes more without the interference of majority classes. In the absence of majority classes, the imbalance ratio between minority classes is much smaller, making it possible to give a more accurate representation. Hence, we propose learning an auxiliary pool $\mathcal{A}^{aux}$, with the same pool size as $\mathcal{A}$, to favor minority classes. Similarly, the auxiliary adapted model gives the prediction
\begin{equation*}
    f(\mathbf{x}; \mathcal{A}^{aux}) = g^{aux}(\phi(\mathbf{x};\mathcal{A}^{aux}))
\end{equation*}
where $\phi(\cdot;\mathcal{A}^{aux})$ and $g^{aux}(\cdot)$ are the corresponding backbone and the classifier. Both classifiers are extended in the way mentioned in section \ref{ltcil}. The retrieval loss based on the auxiliary pool is dubbed as $\mathcal{L}(\mathbf{x}, y; \mathcal{A}^{aux})$.
In this case, the original pool will be trained for all classes, while the auxiliary pool should be specially optimized on tail classes, enabling a holistic representation among all classes.
This can be realized by adjusting the weights of the auxiliary loss for different classes. The weight of the loss is determined by the frequency of the class in the training set:

\begin{equation}
\label{origin}
\mathcal{L}_{1}(\mathbf{x}, y) = \mathcal{L}(\mathbf{x}, y; \mathcal{A}) + \mathbb I(N(y) \le \theta) \mathcal{L}(\mathbf{x}, y; \mathcal{A}^{aux}) \,,
\end{equation}
where $N({y})$ is the number of instances belonging to class $y$, $\mathbb I(\cdot)$ is the indicator function, which outputs $1$ if the expression holds and 0 otherwise. $\theta$ is the threshold to define minority and majority classes. For classes with instances less than $\theta$, we consider it as a minority class and train the auxiliary pool to fit features for them.

\noindent\textbf{Effect of Auxiliary Adapter Pool:}  Eq. \ref{origin} introduces the auxiliary pool and sums up losses, which is shown in the middle part of Figure \ref{figure:method}. The first item forces adapters to learn from current task and update stored knowledge, and the second forces the auxiliary pool to learn from minority classes only. By optimizing Eq. \ref{origin}, on the basis of sufficient learning of majority classes, the auxiliary pool generalizes on minority classes. As a result, the bias in the recognition of minority and forgetting in continual learning is alleviated. 

\subsection{Adaptive Routing}

The weight function $\mathbb I(N(y) \le \theta) $ reflects the correlation between the auxiliary pool and all classes, especially the minority classes. The auxiliary pool does not retain valuable information from the majority, only from the minority, as a modification of one single pool. However, the heuristic format of the step function relies on $\theta$, making a hard and artificial boundary between majority classes and minority classes. Filtering needs a precise $\theta$. For some long-tailed distributions with an extreme imbalance ratio, the instances of one majority class may be nearly as much as the sum of the instances of all minority classes in a task. In this case, a considerable $\theta$ is needed. Once $\theta$ gets smaller, unexpected data may be excluded, leaving them insufficient representations even with an auxiliary pool. Besides, the step function makes no difference between minority classes, meaning the imbalance between minority classes is still unsettled.

To reduce dependency on a precise threshold and modify the minority imbalance, we propose an adaptive adapter routing strategy to assign samples with an instance-specific function for a smoother boundary. The information of the instance and the category are combined to represent the relation to $\mathcal{A}^{aux}$. For instance $\mathbf{x}$, the instance information is encoded in the embedding of $\phi(\mathbf{x})$ with a linear layer $\psi_1(\cdot)$, \emph{i.e.}, $\psi_1(\phi(\mathbf{x}))$, and the class information is encoded with a mapping from integers to embeddings, \emph{i.e.}, $\psi_2(N({y}))$. The concatenation of two embeddings is passed through an $\textbf{MLP}$ to get the adaptive weight:
\begin{equation}
\label{assigner}
\begin{aligned}
w(\mathbf{x},y) = \sigma(\textbf{MLP}([\ \psi_1(\phi(\mathbf{x})), \psi_2(N({y}))])) \,.
\end{aligned}
\end{equation}
We train an assigner $w$ for each instance. The output of the assigner is between 0 and 1 after a non-linear activation $\sigma(\cdot)$, reflecting the correlation between the auxiliary pool and minority classes. Then, the origin loss is updated to 
\begin{equation}
\label{new}
\begin{aligned}
    \mathcal{L}_1(\mathbf{x}, y) = \mathcal{L}(\mathbf{x}, y; \mathcal{A}) +  w(\mathbf{x},y)\mathcal{L}(\mathbf{x}, y; \mathcal{A}^{aux}) \,.
\end{aligned}
\end{equation}

\begin{algorithm}
\caption{Adaptive Adapter Routing for LTCIL}\label{alg1}
{\bf Input}: Dataset: $D_t$. Pre-trained model: $\phi(\cdot)$;\\
{\bf Output}: $\mathcal{A}$, $\mathcal{A}^{aux}$, $w(\cdot,\cdot)$, $g(\cdot)$, $g^{aux}(\cdot)$;
\begin{algorithmic}[1]
            \State Randomly initialize $\mathcal{A}$, $\mathcal{A}^{aux}$;
			\Repeat
			\State Get a mini-batch of training instances: $\{(\mathbf{x}_i,y_i)\}_{i=1}^{n_t}$;
			\State Calculate the loss $\mathcal{L}$ based on $\mathcal{A}$;
            \State Calculate the loss $\mathcal{L}$ based on $\mathcal{A}^{aux}$;
            \State Calculate the weighted sum of loss $\mathcal{L}_1$ in Eq.~\ref{new};
			\State Calculate the regularization term $\mathcal{L}_2$ in Eq.~\ref{eq:reg};
			\State Get the total loss $\mathcal{L}_1+\mathcal{L}_2$; 
			\State Obtain derivative and update the model;
			\Until{reaches predefined epochs}
\end{algorithmic}
\end{algorithm}

\begin{table*}[t]
	\caption{ Average and last performance comparison on three datasets in \textbf{shuffled} LTCIL.  `IN-R' stands for `ImageNet-R' and `ObjNet' stands for `ObjectNet'. The best performance is shown in bold. All methods are implemented with the same pre-trained model for a fair comparison. Methods with $\dagger$ require exemplars while others do not.
	}\label{tab:benchmark}
	\centering
		\resizebox{1.0\textwidth}{!}{%
\begin{tabular}{lcccccccccccc}
\toprule
\multirow{2}{*}{Shuffled LTCIL} & \multicolumn{2}{c}{CIFAR B50-5} & \multicolumn{2}{c}{CIFAR B50-10} & \multicolumn{2}{c}{IN-R B100-10} & \multicolumn{2}{c}{IN-R B100-20} & \multicolumn{2}{c}{ObjNet B100-10} & \multicolumn{2}{c}{ObjNet B100-20} \\
             & $\overline{\text{Acc}}$    &$\text{Acc}_T$    &$\overline{\text{Acc}}$     &$\text{Acc}_T$          &$\overline{\text{Acc}}$       &$\text{Acc}_T$      & $\overline{\text{Acc}}$&$\text{Acc}_T$  & $\overline{\text{Acc}}$     &$\text{Acc}_T$          &$\overline{\text{Acc}}$     &$\text{Acc}_T$                \\ \midrule
    Finetune&60.62 &48.51& 66.79 & 56.45& 67.08&54.73&71.18& 63.75&30.66& 21.08 &35.44&26.40 \\
    LwF&63.48 &48.62 & 70.85 & 60.50&72.78&64.88& 76.56& 70.70&36.88&27.66 &38.27&31.85    \\
    LUCIR$^\dagger$& 79.98& 76.08& 81.59& 78.84&75.23&69.53&77.81& 73.25&45.54& 42.61 &46.69& 43.69         \\
    LUCIR+LWS$^\dagger$& 80.51& 75.90& 82.34& 79.55& 76.44& 71.22 &78.70&75.20  &46.15& 40.46 & 48.13& 45.20          \\
    SimpleCIL&69.81&66.53& 69.97& 66.53& 56.38&54.52&56.55&54.52 &37.75&34.58 &37.57&34.58          \\
    ADAM w/ Finetune&75.22&72.96& 75.42& 72.96 &62.55&61.32&62.64&61.32 &48.13&44.82& 47.96&44.82          \\
    L2P  & 73.59&67.17& 76.04& 71.73&68.93&62.65& 72.85& 68.28&44.15& 39.74 &45.44&42.17          \\
    DualPrompt&69.00  &63.08 & 72.75 & 67.45&69.20&65.07& 71.83& 68.60&42.20& 37.49 &43.95&40.49   \\
    CODA-Prompt&76.45 & 70.29&  79.39& 74.40&75.58&71.82& 78.46& 75.45&46.66& 41.55 &48.81&44.87          \\
    \midrule
    \namea& \textbf{84.91}& \textbf{81.93}& \textbf{86.10}& \textbf{83.89}& \textbf{78.65}&\textbf{75.50}& \textbf{80.16}& \textbf{77.03}&\textbf{53.74}& \textbf{50.59} &\textbf{52.88}&\textbf{48.30}          \\
    \bottomrule               
\end{tabular}
}
\end{table*}

\begin{table*}[t]
	\vspace{-2mm}
	\caption{ Average and last performance comparison on three datasets in \textbf{ordered} LTCIL.  `IN-R' stands for `ImageNet-R' and `ObjNet' stands for `ObjectNet'. The best performance is shown in bold. All methods are implemented with the same pre-trained model for a fair comparison. Methods with $\dagger$ require exemplars while others do not.
	}\label{tab:benchmark2}
	\centering
	\resizebox{1.0\textwidth}{!}{%
\begin{tabular}{lcccccccccccc}
\toprule
\multirow{2}{*}{Ordered LTCIL} & \multicolumn{2}{c}{CIFAR B50-5} & \multicolumn{2}{c}{CIFAR B50-10} & \multicolumn{2}{c}{IN-R B100-10} & \multicolumn{2}{c}{IN-R B100-20} & \multicolumn{2}{c}{ObjNet B100-10} & \multicolumn{2}{c}{ObjNet B100-20} \\
             & $\overline{\text{Acc}}$    &$\text{Acc}_T$    &$\overline{\text{Acc}}$     &$\text{Acc}_T$          &$\overline{\text{Acc}}$       &$\text{Acc}_T$      & $\overline{\text{Acc}}$&$\text{Acc}_T$  & $\overline{\text{Acc}}$     &$\text{Acc}_T$          &$\overline{\text{Acc}}$     &$\text{Acc}_T$                \\ \midrule
    Finetune&    66.06&43.80 &70.39 & 45.73&67.97&48.85&72.13& 60.80&21.52&00.47 &24.73& 10.32         \\
    LwF&70.12&52.33 & 77.14& 64.78&75.53&64.13&78.20&71.97 &23.54& 00.44 &21.78&02.94          \\
    LUCIR$^\dagger$&79.94& 70.73& 83.52&75.73 &77.51&71.28&80.32&75.40 &51.11& 42.43 &50.57& 42.21         \\
    LUCIR+LWS$^\dagger$&80.59&70.44 & 83.24& 75.69&78.03&71.62& 80.46& 75.18& 52.89& 44.57 &53.74& 45.32         \\
    SimpleCIL&72.22&67.67 & 72.36& 67.67&57.39&54.52&57.32& 54.52&44.95& 34.58 &44.72&34.58          \\
    ADAM w/ Finetune&77.10  &72.98 & 77.19& 72.98&62.91&61.30&62.91&61.30 &51.16& \textbf{44.84} &51.16&44.84          \\
    L2P  & 74.68&59.36  & 78.37& 65.48&72.75&65.92&74.95& 69.40&49.94& 40.53 &50.87&41.76          \\
    DualPrompt&74.80& 60.72& 77.88 & 65.84&71.24&66.65&72.85& 69.08&47.88&39.14  &49.15& 40.74         \\
    CODA-Prompt&75.36 &59.07 & 80.44& 67.46&78.33&\textbf{74.42}&79.94& 76.97&51.43&41.42  &52.93& 43.26         \\
    \midrule
    \namea& \textbf{84.21} & \textbf{73.09}& \textbf{87.16}& \textbf{78.90}& \textbf{78.92}& 74.03&\textbf{81.41}&\textbf{ 77.05}&\textbf{55.94}&43.83&\textbf{57.00}&\textbf{46.61}          \\
    \bottomrule                 
\end{tabular}
}
\vspace{-2mm}
\end{table*}

To instruct the routing with more data information, the original weight in the format of the step function can be seen as a reference at the first epochs. Then, we avoid the tendency to 0 of weights when training by adding a regularization term:
\begin{equation} \label{eq:reg}
\begin{aligned} 
    \mathcal{L}_{2}(\mathbf{x},y) = (\alpha-w(\mathbf{x},{y}))^2 \,,
\end{aligned}
\end{equation}
where $\alpha$ is a hyperparameter reflecting the restriction on the optimization of the assigner $w$. We set it to 1 as default.

\noindent\textbf{Effect of Adaptive Adapter Routing:} Eq.~\ref{assigner} learns an instance-wise weight for the auxiliary loss, which is shown in the left part of Figure \ref{figure:method}. It combines the information in the instance and the class. The replacement in Eq. \ref{new} is shown in the right part of Figure \ref{figure:method}. Compared to the heuristic weight, the adaptively learnable weight regulates the importance of a single instance in a data-driven manner. Thus, the auxiliary pool can encode more instance-specific knowledge and give a more comprehensive representation.

\subsection{Summary of \namea}
We give the pseudo code of \name in Algorithm \ref{alg1}. In each mini-batch, we first calculate the loss produced by the adapter pool separately as Eq.~\ref{pool}. Then we learn weight for auxiliary loss following Eq.~\ref{assigner} and sum them up for back propagation, {\em i.e.,} $\mathcal{L}_1(\mathbf{x}, y) + \mathcal{L}_{2}(\mathbf{x}, y)$. Note that the two adapter pools share the same pre-trained model, making the memory budget negligible compared to the PTM.

During inference, we add the logits of two adapter pools by a simple ensemble, {\em i.e.,} $f(\mathbf{x}; \mathcal{A}) + f(\mathbf{x}; \mathcal{A}^{aux})$.
The routing function is only adopted during training, and the logits during inference is the addition of two adapted models without relying on the routing module. The forward pass to obtain \textbf{[CLS]} token is necessary if the key-value mechanism is applied in the PTM-based methods. Thus, compared to those, \name needs only one more forward at inference. One more forward makes a better performance at the cost of expensive inference, which may be improved by simplifying the structure, such as reducing the adapted layers.

\begin{figure*}[t]
	\vspace{-4mm}
	\begin{center}
		\subfigure[CIFAR B50-5]
		{\includegraphics[width=.3\columnwidth]{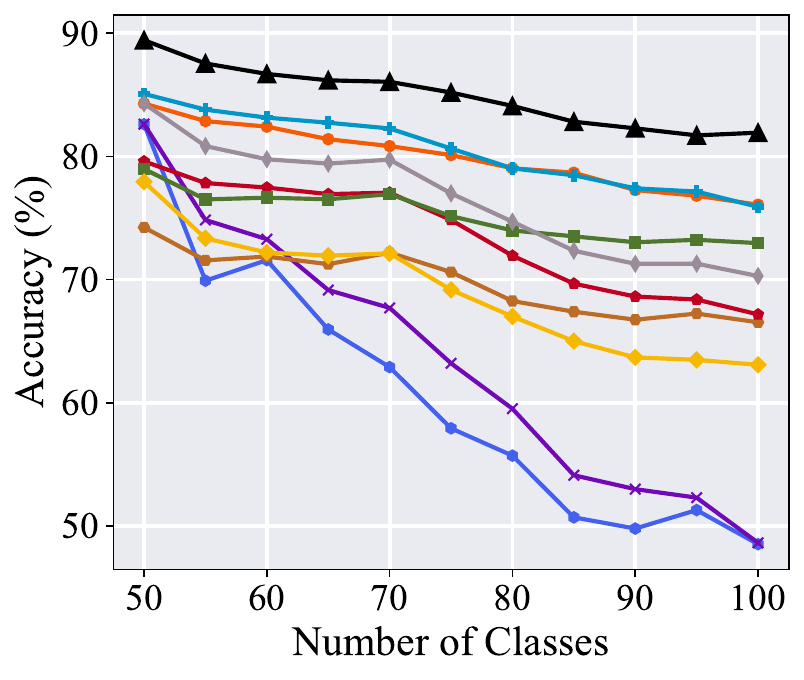}
		\label{figure:vit-cifar}
		}
		\subfigure[ImageNet-R B100-10]
		{\includegraphics[width=.3\columnwidth]{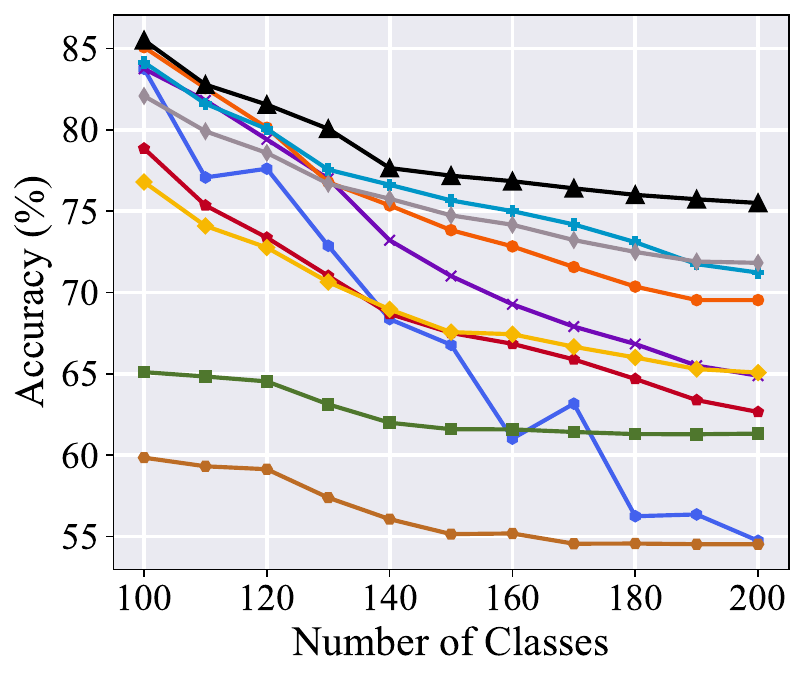}
		\label{figure:vit-cub}
		}
		\subfigure[ObjectNet B100-10]
		{\includegraphics[width=.3\columnwidth]{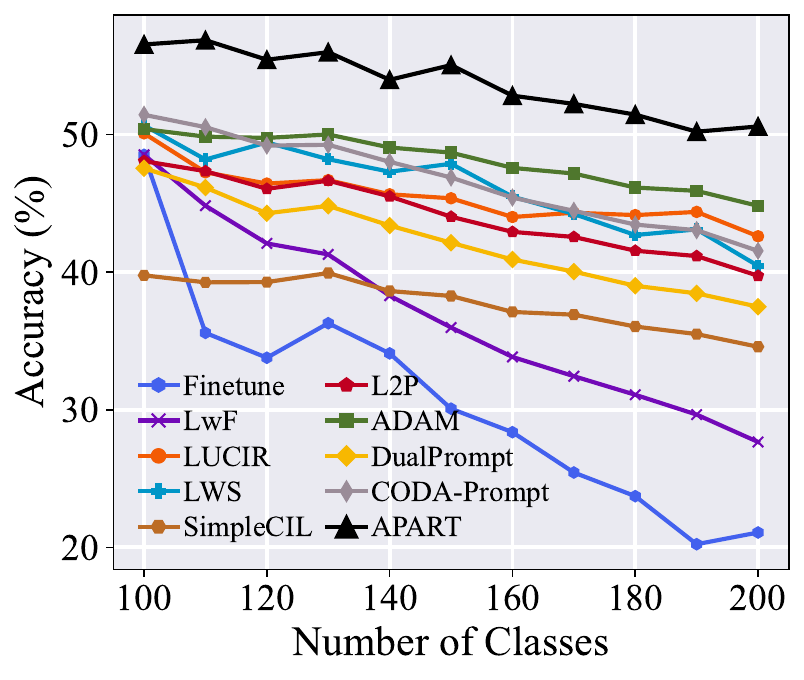}
		\label{figure:vit-ina}
		}	
	\end{center}
	\caption{ Incremental performance when starting from half of the total classes in \textbf{shuffled} LTCIL. We show the legends in \textbf{(c)}. \name consistently outperforms other compared methods.
	} \label{figure:benchmark}
\end{figure*}

\section{Experiment}\label{exp}

In this section, we compare \name on benchmark LTCIL
datasets with state-of-the-art methods. The ablations verify the effectiveness of each part of \namea, and further analysis and visualization are conducted to explore the inherent characteristics of \namea.

\subsection{Implementation Details}

\noindent\textbf{Dataset:} Following~\cite{liu2022long}, we first experiment on the dataset CIFAR100~\cite{krizhevsky2009learning}. Since PTMs are mostly pre-trained on ImageNet21k~\cite{russakovsky2015imagenet}, datasets like ImageNet-Subset with 100 classes are unsuitable for evaluation due to the overlap. Following~\cite{wang2022dualprompt,revistingcil}, we choose another two datasets ImageNet-R~\cite{hendrycks2021many} and ObjectNet~\cite{barbu2019objectnet} as challenging downstream tasks for PTMs to adapt to. Among them, CIFAR100 contains 60,000 pictures for 100 classes. ImageNet-R contains 30,000 pictures for 200 classes. ObjectNet contains about 33,000 pictures for 200 classes. To simulate LTCIL scenarios, we sample part of these datasets. Following~\cite{liu2022long}, we control the long-tailed distribution by a parameter $\rho$, which is the ratio between the quantity of the least frequent class $N_{min}$ and that of the most frequent class $N_{max}$, \emph{i.e.}, $\rho = \frac{N_{min}}{N_{max}}$. For CIFAR100, the imbalance ratio $\rho$ is set to 0.01, and, at most, one class has 500 instances. For ImageNet-R, the dataset is naturally long-tailed with $\rho=0.11$ and the most frequent class has 349 instances. Although it does not follow a standard exponential decay, we leave it as origin without sampling. For ObjectNet, we set $\rho=0.01$ and $N_{max}=200$. 

\noindent\textbf{Setting:} Following~\cite{liu2022long}, we conduct two LTCIL scenarios, {\em i.e.,} ordered LTCIL and shuffled LTCIL. The former follows the long-tailed distribution by task, while the latter first shuffles the decaying numbers randomly and then assigns frequency to each class. In ordered LTCIL, the imbalance ratio remains identical across tasks, while shuffled LTCIL can be seen as a general scenario allowing different ratios in different tasks.

\noindent\textbf{Dataset split:} Firstly, we adopt the split in~\cite{liu2022long} that starts with a base task containing half classes and then separates other classes into 5 tasks or 10 tasks. For simplicity, we denote the split in the format of ``B$\{m\}$-$\{n\}$'', where $m$ is the number of classes in the first task and $n$ is the number in the following tasks. 

\noindent\textbf{Compared methods:} Since our method is based on PTMs, we mainly compare to PTM-based CIL methods, {\em i.e.,} L2P~\cite{l2p}, DualPrompt~\cite{wang2022dualprompt}, CODA-Prompt~\cite{seale2022coda}, SimpleCIL~\cite{revistingcil} and ADAM-Finetune~\cite{revistingcil}. 
Besides, we also re-implement LWS~\cite{liu2022long} with PTMs, which is an exemplar-based method. Since it is a training trick that needs to be combined with other methods, we choose LUCIR~\cite{hou2019learning} and LUCIR+LWS for comparison. Finally, we also compare to the classical CIL algorithm, LwF~\cite{li2017learning}, and the baseline method, Finetune, which saves no exemplars.  

\noindent\textbf{Training details:} Following~\cite{l2p}, we use the same backbone \textbf{ViTB/16-IN1K}, which is pretrained on ImageNet21K and additionally finetuned on ImageNet1K {\em for all compared methods}. The choice of backbone determines the dimension of $\mathbf{k}_s$, as the retrieval is based on the cosine distance between the keys and the embedding. Thus, the dimension of $\mathbf{k}_s$ is set to $768$. We train the model using Adamw with a batch size of 48 for 10 epochs. The learning rate starts from 0.003 and decays with cosine annealing. For \namea, the size of the pool is 5, and the projection dimension is 64. For prompt-based methods, the size of the pool is 10. For exemplar-based methods like LUCIR and LWS, we save 10 exemplars per class for replay. 

\noindent\textbf{Evaluation protocol:} Following~\cite{rebuffi2017icarl}, we record the accuracy after each task $i$ as $\text{Acc}_i$ and use the average $\overline{\text{Acc}}=\frac{1}{T}\sum_{i=1}^T \text{Acc}_i$ on all $T$ tasks and the last accuracy $\text{Acc}_T$ as the metrics.

\subsection{Benchmark Comparison}
In this section, we report the accuracy over benchmark
datasets, {\em i.e.,} CIFAR100, ImageNet-R, and ObjectNet under both shuffled LTCIL and ordered LTCIL in Table \ref{tab:benchmark}, \ref{tab:benchmark2} and show the incremental performance in Figure \ref{figure:benchmark}. 

Specifically, Figure \ref{figure:benchmark} clearly shows the superior performance of \namea. 
We can infer that conventional methods ({\emph{e.g.}, LwF}) have an evident downward trend, showing they suffer from catastrophic forgetting. In contrast, the decline of LUCIR and LWS is relatively modest, owing to the help of exemplars. Compared to other parameter-efficient finetuning-based techniques ({\em e.g.,} L2P, DualPrompt and CODA-Prompt), we find a gap between the first task's performance. 
The gap clearly indicates the effectiveness of utilizing auxillinary adapter pool to compensate for the minority classes during training.

Additionally, \name demonstrates its superiority over compared methods on benchmark datasets in Table \ref{tab:benchmark}, \ref{tab:benchmark2}. The poor performance of conventional CIL methods, \emph{i.e.,} LwF, indicates that the long-tailed data magnifies the difficulty of the CIL problem, even based on PTMs with strong generalizability. The representation ability is proved by the performance of SimpleCIL, which totally relies on the frozen PTMs. However, the benefit from PTMs is limited in LTCIL, requiring more task-specific features for improvement. Similarly, \name outperforms most prompt-based methods by even 8\% in the shuffled scenario and 9\% in the ordered scenario. It reveals that, although prompts help resist forgetting in conventional CIL, the knowledge stored in prompts is unavoidably interfered with by imbalance. LWS is designed for convolution networks in LTCIL and improves the performance of combined LUCIR. However, as reported, when we replace ResNet with pre-trained ViT, the improvement of a re-weighting classification layer is limited. Although with a balanced dataset, LWS remains the biased representation unsettled. It shows that, in the long-tailed data stream, the bias in representation is more harmful than the bias in classification. Thus, the auxiliary pool in \name learns more from minority classes, removing the underlying bias of representation, receiving a better performance than LUCIR and LUCIR+LWS, indicating the superiority of a dedicated auxiliary pool for minority classes. To sum up, \name outperforms both prompt- and exemplar-based methods, validating its effectiveness.

\subsection{Ablation Study}

\begin{wraptable}{r}{8.5cm}
\centering
\vspace{-3mm}

\caption{ Ablation study on ordered CIFAR100 B50-5. Each part in \name helps to improve the performance.}

 \label{tab:ablation}
\begin{tabular}{lcc}
\toprule
Method & $\overline{\text{Acc}}$ & \small $\text{Acc}_T$ \\ 
\midrule
\name  &  \bf 84.21     &   \bf 73.09    \\ 
\midrule
w/o Adaptive Routing   &   83.04     &   71.32     \\
w/o Auxiliary Pool       &    80.46    &   67.26     \\
w/o Adapter Pool        &  75.98      &   58.64    \\
\bottomrule
\end{tabular}
 \vspace{-5mm}
\end{wraptable}

In this section, we conduct an ablation study to analyze the importance of three components in \name and explore the influence of the number of adapter pools.

In Table~\ref{tab:ablation}, results clearly show the effectiveness of each component. 
When we drop the adaptive routing between two pools, we train two pools as Eq. \ref{origin}, where the distribution of instances largely depends on a fixed threshold. In contrast, the adaptive and dynamic boundary can obtain an improvement of the auxiliary pool. 
Besides, ablating the auxiliary pool means only using one pool for training, resulting in no unique mechanism to cope with the imbalanced data. Hence, we find one pool cannot generate a comprehensive representation for all classes, and insufficient learning from minority classes leads to a decline. Furthermore, we replace the adapter pool with one single adapter. Due to the lack of capacity, the model suffers from forgetting. The decline in average accuracy is up to 4\% and in last accuracy is nearly 9\%, showing the necessity of multiple adapters when learning sequential tasks.

From the above ablations, we find one more adapter pool can lead to performance improvement. However, does that mean more adapter pools will definitely lead to better performance? Table \ref{tab:ablation2} shows the change in accuracy when applying multiple pools. The added pools are used to capture more from minority classes under the same regulation. From the table, we observe that, at first, the additional pool brings an increase in accuracy. The model size grows linearly when the pools increase, but the improvement is limited. When the number of pools increases to 6, a decrease occurs. The results show that the increase in parameters does not necessarily guarantee performance improvement. According to the experiments, we set the size to 2 for a trade-off between performance and model size.

\begin{table}[htbp]
\caption{ Incremental performance with different number of adapter pools on ordered CIFAR100 B50-5.}
\vspace{-3mm}
 \label{tab:ablation2}
\begin{tabular}{ccccccc}
\toprule
\# Pool &  1 &2 &3 &4 &5 & 6 \\ \midrule
$\overline{\text{Acc}}$   & 80.46 & 83.04& 83.11 & 83.40 & 83.28 &83.11 \\
$\text{Acc}_T$ & 67.26 & 71.32 &70.98 &72.13 & 72.16 & 71.75\\
\bottomrule
\end{tabular}
\end{table}

\subsection{Further Analysis}

\noindent\textbf{Subgroup measures:} Following \cite{openlongtailrecognition}, we can draw three splits of all classes by the number of instances and report the performance on these splits. Specifically, we report the accuracy of three splits of classes in CIFAR100 in Table \ref{tab:mmf}, {\em i.e.,} many-shot(with $\ge$ 100 instances), medium-shot (20 $\sim$ 100 instances) and few-shot ($\le$ 20 instances). `Overall' denotes $\text{Acc}_T$. Specifically, we find the poor overall performance of other methods is mainly due to the performance gap in minority classes. As the instances become fewer, test accuracy becomes lower. 
When comparing SimpleCIL to L2P, we find L2P gets a better overall performance at the cost of minority classes. By finetuning the PTMs, L2P increases the many-shot accuracy by 13\% but improves overall accuracy by 0.6\%. The bias in representation brings the neglect of minority classes. By contrast, \name obtains a holistic improvement on different class sets.\\

\begin{table}[htbp]
\caption{Group Accuracy on shuffled CIFAR100 B50-5. \name presents a holistic excellence on different classes.}
\vspace{-4mm}
 \label{tab:mmf}
\begin{tabular}{lcccc}
\toprule
Method & Overall & Many & Median& Few \\ \midrule
Finetune & 48.51 &63.80& 51.66& 27.00 \\
LwF & 48.62	& 75.11 & 47.77 & 18.70\\
LUCIR & 76.08	&81.34 & 74.00& 72.37\\
LUCIR+LWS & 75.90 &81.51 & 74.46& 71.03 \\
SimpleCIL& 66.53&68.14 & 64.34 & 67.20\\
ADAM & 72.96 &78.37 & 73.54 & 65.97\\
L2P& 67.17 &81.14 & 62.37 & 56.47\\
DualPrompt& 63.08&78.71 & 58.29& 50.43 \\
CODA-Prompt&70.29 &84.49& 65.63& 59.17\\ \midrule
\namea& \bf 81.93	&\bf89.37 & \bf81.00&\bf74.33 \\
\bottomrule
\end{tabular}
\vspace{-4mm}
\end{table}

\begin{figure}[htbp]
	\begin{center}
		\subfigure[CIFAR B50-5]
		{\includegraphics[width=.35\columnwidth]{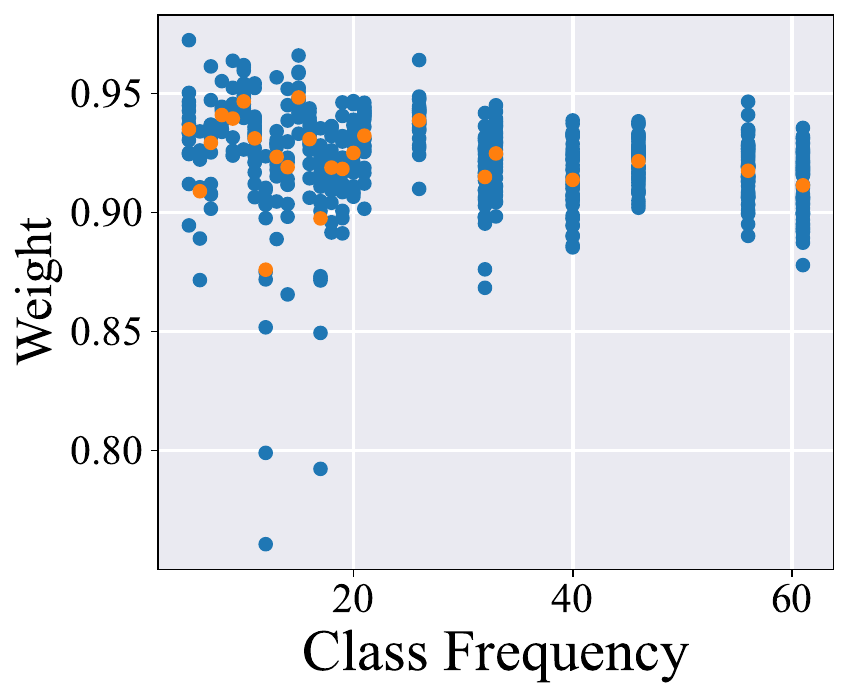}
		}
		\subfigure[ObjectNet B100-10]
		{\includegraphics[width=.35\columnwidth]{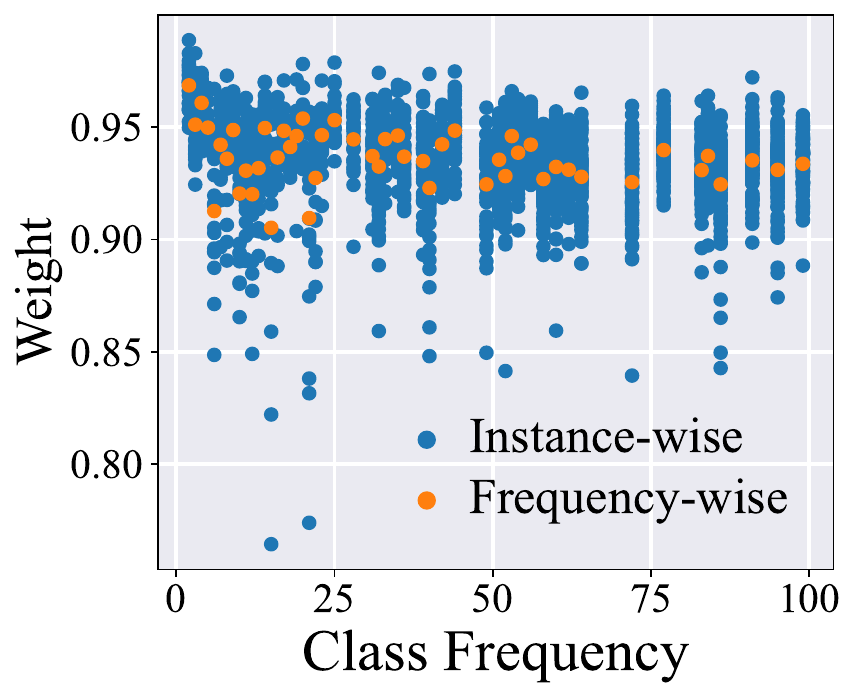}
		}
	\end{center}
	\caption{Adaptively learned weights showing in the view of instance and frequency in \textbf{shuffled} LTCIL. We show the legends in \textbf{(b)}. Weights show a diversity among classes and a decrease with the increase in frequency. 
	} \label{figure:assigner}
\end{figure}

\noindent\textbf{Visualizations of the assigner weights:} In this section, we show the weight learned by the assigner in Figure \ref{figure:assigner}, {\em i.e.,} Eq.~\ref{assigner}. It reveals the relationship between learned weight $w(\mathbf{x},y)$ and class frequency $N(y)$ (mentioned in Eq.~\ref{origin}) for each instance. `Instance-wise' reports the weight of one instance, and `Frequency-wise' reports the average weights with the same class frequency. From the figure, different classes yield diverse weights, revealing that the learned routing encodes task-specific information in a data-driven manner. Besides, for classes with more instances, the learned weights exhibit an overall {\em decreasing} trend. The result implies the adaptive weights help the auxiliary pool learn more from minority classes. Hence, it strengthens the correlation between the auxiliary pool and minority classes and modifies the representation. \\

\begin{table}[htbp]
\small
\vspace{-4mm}
\caption{\centering Fair comparison on ImageNet-R.}
\label{tab:runs}
\vspace{-2mm}
\centering
\begin{tabular}{lcccc}
\toprule
& \multicolumn{2}{c}{Ordered B100-10} & \multicolumn{2}{c}{Shuffled B100-20} \\
& $\overline{\text{Acc}}$    &$\text{Acc}_T$    &$\overline{\text{Acc}}$     &$\text{Acc}_T$  \\ \midrule
iCaRL-352 &  68.95 & 57.22 & 70.24 & 58.52  \\
iCaRL-2000 &    77.43  &  70.08 & 78.11 &  73.12 \\
\name      &  \textbf{78.92} & \textbf{74.03} & \textbf{80.16} & \textbf{77.03} \\ \bottomrule
\end{tabular}
 \vspace{-5mm}
\end{table}

\noindent\textbf{Fair comparison:} \name saves no exemplars but needs tuning parameters. Since the memory consists of parameters and exemplars, we conduct a fair comparison given the same memory budget, as in ~\cite{zhou2022memo}. One rehearsal-based method re-implemented with the same backbone \textbf{ViTB/16-IN1K}, iCaRL~\cite{rebuffi2017icarl}, is compared. Even based on exemplars, iCaRL implemented with randomly initialized ResNet is weaker. To align the memory space, we calculate the corresponding exemplar memory for iCaRL. The memory of \name consists of three parts: one frozen backbone to get \textbf{[CLS]} token, one backbone to be finetuned, and method-related parameters. For iCaRL, the memory consists of three parts: one old backbone for knowledge distillation, one backbone to train, and the exemplar memory. The memory of the exemplar should be consistent with that of method-related parameters, including pools, embeddings, and classifiers. Saving an ImageNet-R image costs $3\times224\times224$ integer numbers (int), while \name costs $12,231,953$ method-related parameters (float). To align the memory budget, iCaRL needs to save $12,231,953$ floats $\times 4$ bytes/float $\div \ (3 \times 224 \times 224)$ bytes/image $ \approx 352$ instances for ImageNet-R. The comparison in Table \ref{tab:runs} shows \name achieves a better performance given the same memory space as iCaRL-352, which stores 352 exemplars. Even with more exemplars up to 2000, iCaRL cannot beat \namea. Exemplars cost more, according to the results.

\section{Conclusion}\label{conclusion}

In our dynamic world, data often comes in an imbalanced streaming manner, requiring the model to tackle long-tailed class-incremental learning. This paper proposes \name for LTCIL, which learns adapter pools for pre-trained models to overcome forgetting via instance-specific selection. To compensate for the tail classes, we learn an auxiliary adapter pool for a unified feature representation. Furthermore, we design an adaptive adapter routing strategy to automatically select the proper pool to use, in order to trace the long-tailed distribution in a data-driven manner. Extensive experiments verify \namea's excellent performance. 


\backmatter

\section*{Acknowledgment}
	This work is partially supported by National Science and Technology Major Project (2022ZD0114805), Fundamental Research Funds for the Central Universities (2024300373),
NSFC (62376118, 62006112, 62250069, 61921006), Collaborative Innovation Center of Novel Software
Technology and Industrialization.

\begin{appendices}

\section{Implementation Details}\label{secA1}

In this section, we discuss the compared methods and details about the benchmark datasets.
\subsection{Compared Methods}

The compared methods in the main paper are as follows:
\begin{itemize}
    \item \textbf{Finetune:} trains the model directly with new datasets without addressing catastrophic forgetting;
    \item \textbf{LwF:}~\cite{li2017learning} utilizes knowledge distillation to transfer knowledge from the old frozen model to the new model while finetuning new tasks;
    \item \textbf{LUCIR:}~\cite{hou2019learning} is an exemplar-based method, which learns normalized cosine classifiers distinct from the old ones as an improvement of LwF;
    \item \textbf{LWS:}~\cite{liu2022long} is a state-of-the-art exemplar-based LTCIL method. It re-trains the linear classifier with balanced data sampled from reserved exemplars and new task data in a two-stage framework. It can be combined with other exemplar-based methods;
    \item \textbf{L2P:}~\cite{l2p} is a state-of-the-art prompt-based CIL method. It freezes the pre-trained model and adds prompts to adapt to new tasks. A prompt pool is built in ``key-value" pairs. When updating, the most suitable prompts are retrieved from the pool by matching the instance and the keys;
    \item \textbf{DualPrompt:}~\cite{wang2022dualprompt} is a state-of-the-art prompt-based CIL method. It extends the prepended prompts in L2P to prompts inserted into each layer, called general prompts and expert prompts. The former learns knowledge across tasks, and the latter follows the retrieval strategy to learn task-specific knowledge;
    \item \textbf{CODA-Prompt:}~\cite{seale2022coda} is a state-of-the-art prompt-based CIL method. It decomposes the prompts in a weighted-sum format and introduces a learnable attention mechanism to prompt matching;
    \item \textbf{SimpleCIL:}~\cite{revistingcil} is a state-of-the-art PTM-based CIL method. It sets the prototype features extracted from the frozen PTMs as the classifiers without extra training on downstream tasks;
    \item \textbf{ADAM:}~\cite{revistingcil} is a state-of-the-art PTM-based CIL method. It adapts to downstream tasks by efficiently tuning on the first task and merges with the origin frozen model by extracting concatenated prototype classifiers.

\end{itemize}

All these methods are implemented with the same backbone, ViT-B/16-IN1K. 
\renewcommand\thefigure{\arabic{figure}}    
\setcounter{figure}{3} 
\renewcommand\thetable{\arabic{table}}    
\setcounter{table}{5} 
\subsection{Datasets} \label{data}

\begin{figure*}[htbp]
	\vspace{-4mm}
        
	\begin{center}
		\subfigure[CIFAR]
		{\includegraphics[width=.3\columnwidth]{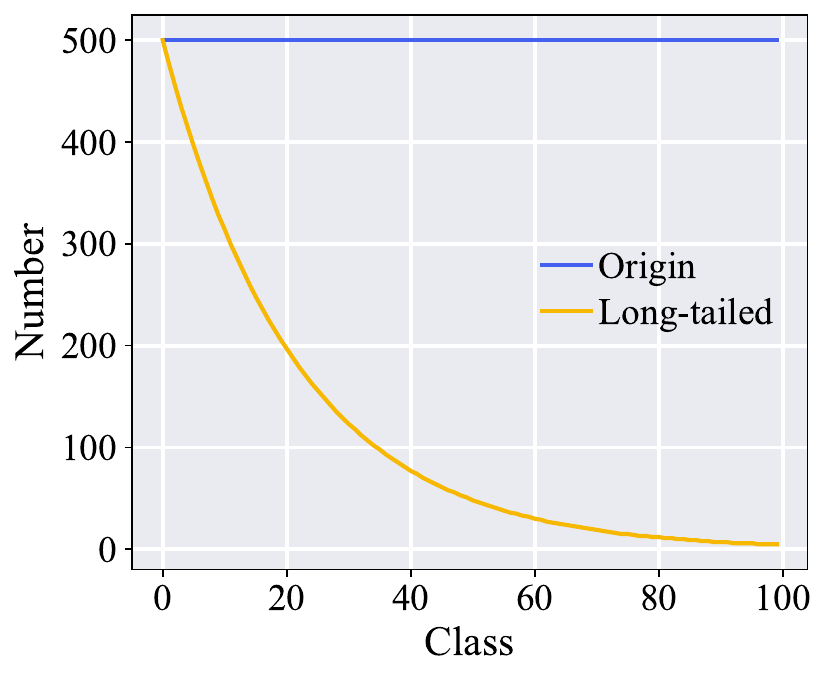}
		
		}
		\subfigure[ImageNet-R]
		{\includegraphics[width=.3\columnwidth]{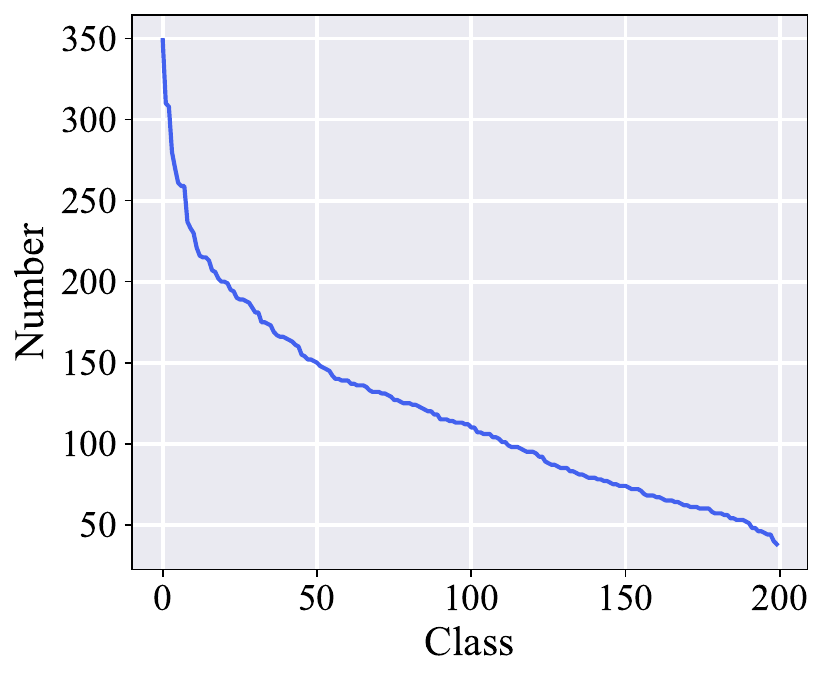}
		
		}
		\subfigure[ObjectNet]
		{\includegraphics[width=.3\columnwidth]{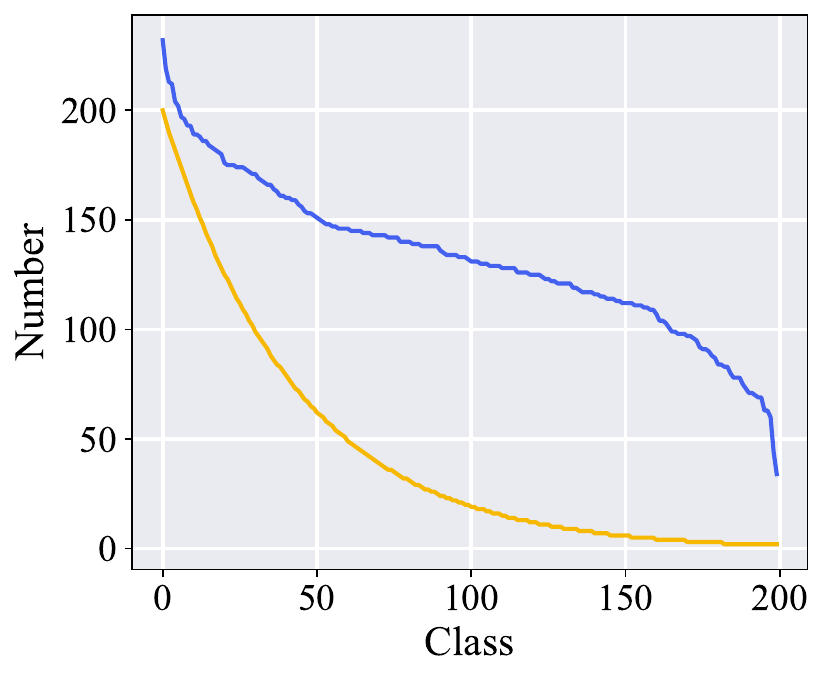}
		
		}	
	\end{center}
	\caption{ Origin distribution and long-tailed distribution after sampling for each dataset. We show the legends in \textbf{(a)}. `Origin' and `Long-tailed' denote the distribution before and after sampling separately.
 \label{figure:dset}
	}
\end{figure*}

Following ~\cite{revistingcil}, we select three datasets for comparison based on pre-trained models. To simulate a long-tailed distribution, we sample from the origin dataset in an exponential decay parameterized by $\rho$~\cite{liu2022long}. $\rho$ is the ratio between the number of the least frequent class and that of the most frequent class, \emph{i.e.}, $\rho = \frac{N_{min}}{N_{max}}$. The detailed introduction is below.

\begin{itemize}
    \item \textbf{CIFAR100:}~\cite{krizhevsky2009learning} contains 60,000 images for 100 classes, of which 50,000 are training instances and 10,000 are testing ones with a uniform distribution. We sample it with $\rho$ = 0.01 and $N_{max}$ = 500;
    \item \textbf{ImageNet-R:}~\cite{hendrycks2021many} is introduced into CIL by ~\cite{wang2022dualprompt}. It contains 30,000 pictures of different styles, of which 24,000 are training instances and 6,000 are testing ones. Since it follows a long-tailed distribution with $\rho$ = 0.11 and $N_{max}$ = 349 for 200 classes, we experiment on it without extra processing;
    \item \textbf{ObjectNet:}~\cite{barbu2019objectnet} is introduced into CIL by ~\cite{revistingcil}. It contains pictures with controlled variations. A subset of 200 classes with about 32,000 instances is selected for evaluation, of which 26,509 are training instances and 6,628 are testing ones. We sample it with $\rho$ = 0.01 and $N_{max}$ = 200.
\end{itemize}

As we sample from the datasets to simulate long-tailed distribution, we provide the distribution in Figure \ref{figure:dset}.

\section{Extra Experimental Evaluations}\label{secA2}
In this section, we analyze more hyperparameters of \namea, besides the number of pools in the main paper. 

\subsection{Influence of Pool Size}

We show the result with different pool sizes in Table ~\ref{tab:adapter}. Pool size $M$ means the number of options in a pool. It reveals that the pool size has an influence on incremental performance. A relatively low accuracy presents when we start from 3 options in a pool, possibly less than the capacity of the model needed for a long sequence of tasks. When a pool has more adapters, an increase occurs, followed by a decrease. The increase is for the proper capacity, while the insufficient learning of each element in the pool may cause the decline. Thus, we set $M$ to 5 as default.

\begin{table}[htbp]
\caption{ Incremental performance with different pool sizes on shuffled CIFAR100 B50-5.}
\vspace{-4mm}
 \label{tab:adapter}
\begin{tabular}{ccccc}
\toprule
 $M$  &3 &5 &7 & 10 \\ \midrule
$\overline{\text{Acc}}$   & 83.61& 84.91& 83.54 & 83.57    \\
$\text{Acc}_T$ & 79.92 & 81.93 & 79.87 & 80.17 \\
\bottomrule
\end{tabular}
 \vspace{-4mm}
\end{table}

\subsection{Influence of Weight Scale}

We explore the influence of the parameter $\alpha$ in $\mathcal{L}_2(\cdot, \cdot)$ which controls the scale of adaptive weights. The choice of $\alpha$ has a significant impact on the performance. The result in Table ~\ref{tab:alpha} shows that the performance improves as $\alpha$ gets larger initially. A small weight directly causes insufficient learning of the auxiliary pool compared to the other pool. Then, the ensemble of predictions is biased when without constraints on it. Meanwhile, the small learned weight makes little gap in routing between majority classes and minority classes, resulting in insufficient learning of minority classes. When $\alpha$ gets larger, a modest decrease occurs, for narrowing the gap similarly. Thus, we set $\alpha$ to 1 as default.

\begin{table}[ht]
\caption{ Incremental performance with different $\alpha$ on shuffled CIFAR100 B50-5.}
\vspace{-4mm}
 \label{tab:alpha}
\begin{tabular}{ccccccc }
\toprule
$\alpha$  & 0.1 &0.3 &0.5 &0.7 & 1.0 & 3.0    \\ \midrule
$\overline{\text{Acc}}$  & 52.58 & 69.74  & 76.35  &  84.02 &  84.91  & 84.40  \\
$\text{Acc}_T$           & 52.61 & 72.28  & 74.57  &  80.84 &  81.93  & 81.51  \\
\bottomrule
\end{tabular}
 \vspace{-4mm}
\end{table}

\end{appendices}

\bibliographystyle{sn-chicago}
\bibliography{acml}

\end{document}